\documentclass{article} 
\usepackage{iclr2019_conference,times}
\usepackage{natbib}

\usepackage[pagebackref=false,breaklinks=false,colorlinks,citecolor=blue,linkcolor=blue]{hyperref}
\usepackage{amsmath}
\usepackage{amsfonts}
\usepackage{amssymb}
\usepackage{url}
\usepackage{bm}
\usepackage{mathtools}
\usepackage{color}
\usepackage{wrapfig}
\usepackage{dsfont}
\usepackage{graphicx,booktabs,tabulary,multirow,overpic,xcolor}
\usepackage[Algorithm]{algorithm}
\usepackage{algpseudocode}
\usepackage{longtable}
\usepackage{colortbl}
\usepackage[caption=false]{subfig}
\usepackage{pifont}
\usepackage{wrapfig}
\usepackage{caption}

\definecolor{Gray}{gray}{0.9}
\newcommand{\tablestyle}[2]{\setlength{\tabcolsep}{#1}\renewcommand{\arraystretch}{#2}\centering\footnotesize}
\newlength\savewidth\newcommand\shline{\noalign{\global\savewidth\arrayrulewidth
		\global\arrayrulewidth 1pt}\hline\noalign{\global\arrayrulewidth\savewidth}}

\newcolumntype{x}[1]{>{\centering\arraybackslash}p{#1pt}}

\newcommand{\cmark}{\ding{51}}%
\newcommand{\xmark}{\ding{55}}%

\newcolumntype{x}[1]{>{\centering\arraybackslash}p{#1pt}}
\newcommand{\dt}[1]{\fontsize{8pt}{.1em}\selectfont \emph{#1}}
\makeatletter\renewcommand\paragraph{\@startsection{paragraph}{4}{\z@}
	{.5em \@plus1ex \@minus.2ex}{-.5em}{\normalfont\normalsize\bfseries}}\makeatother

\title{Feature Intertwiner for Object Detection}

\author{Hongyang Li$^{1}$, Bo Dai$^{1}$, Shaoshuai Shi$^{1}$, Wanli Ouyang$^{2}$ \& Xiaogang Wang$^{1}$
\\[1.5mm]
$^{1}$
Multimedia-SenseTime Joint Lab, The Chinese University of Hong Kong\\
\texttt{yangli@ee.cuhk.edu.hk},~~\texttt{bdai@ie.cuhk.edu.hk},~~\texttt{\{ssshi,xgwang\}@ee.cuhk.edu.hk}
\\[1mm]
$^{2}$ SenseTime Computer Vision Research Group, The University of Sydney\\
\texttt{wanli.ouyang@sydney.edu.au}
}
 
\iclrfinalcopy 
\begin{document}

\maketitle

\begin{abstract}
	A well-trained model should classify objects with a unanimous score for every category. This requires the high-level semantic features should be as much alike as possible among samples.
	%
	To achive this, previous works focus on re-designing the loss 
	or proposing new regularization constraints.
	In this paper, we provide a new perspective.
	For each category, it is assumed that there are two feature sets:
	one with reliable information and the other with less reliable source.
	We argue that the reliable set could guide the feature learning of the less reliable set during training - in spirit of student mimicking teacher's behavior and thus pushing towards a more compact class centroid in the feature space. 
	Such a scheme also benefits the reliable set since samples become closer within the same category - implying that it is easier for the classifier to identify. 
	%
	We refer to this mutual learning process as \textit{feature intertwiner} and embed it into object detection.
	%
	It is well-known that objects of low resolution are more difficult to detect due to the loss of detailed information during network forward pass (\textit{e.g.}, RoI operation). We thus regard objects of high resolution as the reliable set and objects of low resolution as the less reliable set. 
	Specifically, an intertwiner is designed to minimize the distribution divergence between two sets. 
	%
	%
	The choice of generating an effective feature representation for the reliable set 
	is further investigated, where  
	we introduce the optimal transport (OT) theory into the framework. Samples in the less reliable set are better aligned 
	with aid of OT metric. 
	%
	Incorporated with such a plug-and-play intertwiner, we achieve an evident improvement over previous state-of-the-arts.
\end{abstract}

\section{Introduction}\label{sec:introduction}

Classifying complex data in the high-dimensional feature space is the core of most machine learning problems, especially with the emergence of deep learning for better feature embedding \citep{krizhevsky12_alexnet,he2016_resnet,li2018_capsule,li2019_cmt,guo2018learning} . 
Previous methods address the feature representation problem by the conventional cross-entropy loss, $l_1$ / $l_2$ loss, or a regularization constraint on the loss term to ensure small intra-class variation and large inter-class distance \citep{janocha2017_loss,liu2017_coco_v2,wen2016_center_loss,liu2017_spherical_loss}. The goal of these works is to learn more compact representation for each class in the feature space. 
In this paper, we also aim for such a goal and propose a new perspective to address the problem. 

Our observation is that samples can be grouped into two sets in the feature space. One set is more reliable, while the other is less reliable. For example, visual samples may be less reliable due to low resolution, occlusion, adverse lighting, noise, blur, \textit{etc.} The learned features for samples from the reliable set are easier to classify than those from the less reliable one. Our hypothesis is that the reliable set can guide the feature learning of the less reliable set, in the spirit of a teacher supervising the student. 
We refer to this mutual learning process as a feature intertwiner.


%

In this paper, a plug-and-play module, namely, feature intertwiner, is applied for object detection, which is the task of classifying and localizing objects in the wild. An object of lower resolution will inevitably lose detailed information during the forward pass in the network. Therefore, it is well-known that the detection accuracy drops significantly as resolutions of objects decrease.
We can treat samples with high resolution (often corresponds to large objects or region proposals) as the reliable set and samples with low resolution (small instances) as the less reliable set\footnote{We use the term `large object/(more) reliable/high resolution set' interchangeably in the following to refer to the same meaning; likewise for the term `small set/less reliable set/low-resolution set'.}.
Equipped with these two `prototypical' sets, we can apply the feature intertwiner where the reliable set is leveraged to help the feature learning of the less reliable set.

\begin{figure}
	\begin{minipage}[c]{0.7\textwidth}
		\includegraphics[width=\textwidth]{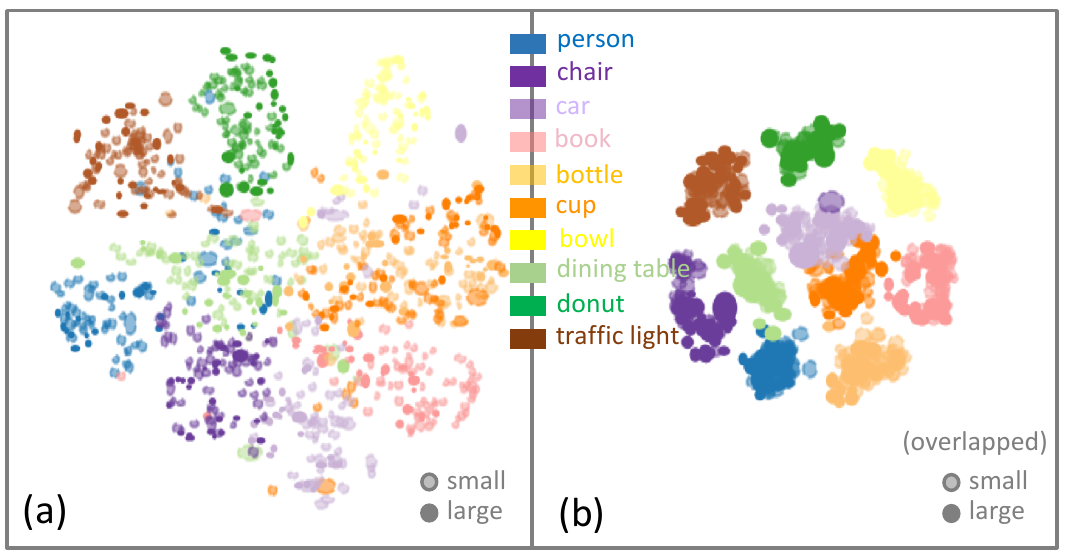}
	\end{minipage}\hfill
	\begin{minipage}[l]{0.28\textwidth}
	\caption{
		(Zoom in for better view) Visualization of the features in object detection using t-SNE \citep{vanDerMaaten2008_tsne} (a) without and (b) with feature intertwiner on COCO.
		Each point is a sample mapped onto the low-dim manifold.
	}
	\label{fig:tsne}
\end{minipage}
\end{figure}
Fig. \ref{fig:tsne} on the left visualizes the learned detection features before classifier\footnote{Only top ten categories with the most number of instances in prediction is visualized. 
For each category, the high-resolution objects (reliable set) are shown in  \texttt{solid} color while the low-resolution instances (less reliable set) are shown in \texttt{transparent} color with dashed boundary.}. 
Without intertwiner in (a), samples are more scattered and separated from each other. Note there are several samples that are far from its own class and close to the samples in other categories (\textit{e.g.}, class \texttt{person} in blue), indicating a potential mistake in classification.
With the aid of feature intertwiner in (b), 
there is barely outlier sample outside each cluster.
the features in the lower resolution set approach closer to the features in the higher resolution set - achieving the goal of compact centroids in the feature space. Empirically, these two settings correspond to the baseline and intertwiner experiments (marked in gray) in Table \ref{tab:ablation:vs_baseline}. The overall mAP metric increases from 32.8\% to 35.2\%, with an evident improvement of 2.6\% for small instances and a satisfying increase of 0.8\% for large counterparts. This suggests the proposed feature intertwiner could benefit both sets.

%

%
%
%
%

Two important modifications are incorporated based on the preliminary intertwiner framework. %
The first is the use of class-dependent historical representative stored in a buffer. Since there might be no large sample for the same category in one mini-batch during training,
the record of all previous features of a given category for large instances is recorded by a representative, of which value gets updated dynamically as training evolves.
The second is an inclusion of the optimal transport (OT) divergence as a deluxe regularization in the feature intertwiner. 
OT metric maps the comparison of two distributions on high-dimensional feature space onto a lower dimension space so that it is more sensible to measure the similarity between two distributions.
For the feature intertwiner, OT is capable of enforcing the less reliable set to be better aligned with the reliable set. 

%
We name the detection system equipped with the feature intertwiner as \textbf{InterNet}.  Full code suite is available at \href{https://github.com/hli2020/feature_intertwiner}{
\texttt{\textcolor{blue}{https://github.com/hli2020/feature\_intertwiner}}}. For brevity, we put the descriptions of dividing two sets in the detection task, related work (partial), background knowledge on OT theory and additional experiments in the appendix.

\section{Related Work}\label{sec:related-work}

\textbf{Object detection}  
\citep{dai2016_rfcn,lin2017_FPN,redmon2016_yolo_v2,li2018_gradient,lu2018_grid,shi2018_pointrcnn}
is one of the most fundamental computer vision tasks and serves as a precursor step for other high-level  problems. 
It is challenging due to the complexity of features in high-dimensional space 
\citep{krizhevsky12_alexnet}, the large intra-class variation and inter-class similarity across categories in benchmarks
\citep{imagenet_conf,coco}. 
Thanks to the development of deep networks structure \citep{simonyan2015_vgg,he2016_resnet} and modern GPU hardware acceleration, this community has witnessed
a great bloom in both performance and efficiency. 
\textbf{The detection of small objects} is addressed in concurrent literature mainly through two manners. The first is by looking at the surrounding context \citep{li16_attentive_context,mottaghi14_context} since a larger receptive filed in the surrounding region 
could well compensate for the information loss on a tiny instance during down-sampling in the network. 
%
%
The second  is to adopt a multi-scale strategy \citep{li2018_zoom_journal,lin2017_FPN,liu2015_ssd,shrivastava2016_top_down_modulation} to handle the scale problem. This is probably the most effective manner to identify objects in various sizes and can be seen in (almost)
all detectors. Such a practice is a ``sliding-window'' version of warping features across different stages in the network, aiming for normalizing the sizes of features for objects of different resolutions.
The proposed feature intertwiner is perpendicular to these two solutions. We provide a new perspective of addressing the detection of small objects - leveraging the feature guidance from high-resolution reliable samples. 

\textbf{Designing loss functions for learning better features.} 
The standard cross-entropy loss does not have the constraint on narrowing down the intra-class variation. 
Several works thereafter have focused on adding new constraints to the intra-class regularization.
Liu \textit{et al}. \citep{liu2017_spherical_loss} proposed the angular softmax loss to learn angularly discriminative features. The new loss is expected to have smaller maximal intra-class distance than
minimal inter-class distance. 
The center loss \citep{wen2016_center_loss} approach specifically learns a centroid for each class and penalizes the distances between samples within the category and the center. Our feature intertwiner shares some spirit with this work in that, the proposed buffer is also in charge of collecting feature representatives for each class. A simple modification \citep{liu2017_coco_v2} 
to the inner product between the normalized feature input and the class centroid for the softmax loss also decreases the inner-class variation and improves the classification accuracy. Our work is from a new perspective in using the reliable set for guiding the less reliable set.
 

\section{Feature Intertwiners for Object Detection}


In this paper, we adopt the Faster RCNN pipeline for object detection \citep{he2016_resnet,he2017_mask_rcnn,ross15_fast_rcnn}. In Faster RCNN, the input image is first fed into a backbone network to extract features; a region proposal network \citep{ren2015_faster_rcnn} is built on top of it to generate potential region proposals, which are several candidate rectangular boxes that might contain objects. These region proposals vary in size. Then the features inside the region are extracted and warped into the same spatial size (by RoI-pooling). Finally, the warped features are used by the subsequent CNN layers for classifying whether an object exists in the region. 



\begin{figure}[h]
\centering
		\includegraphics[width=.8\textwidth]{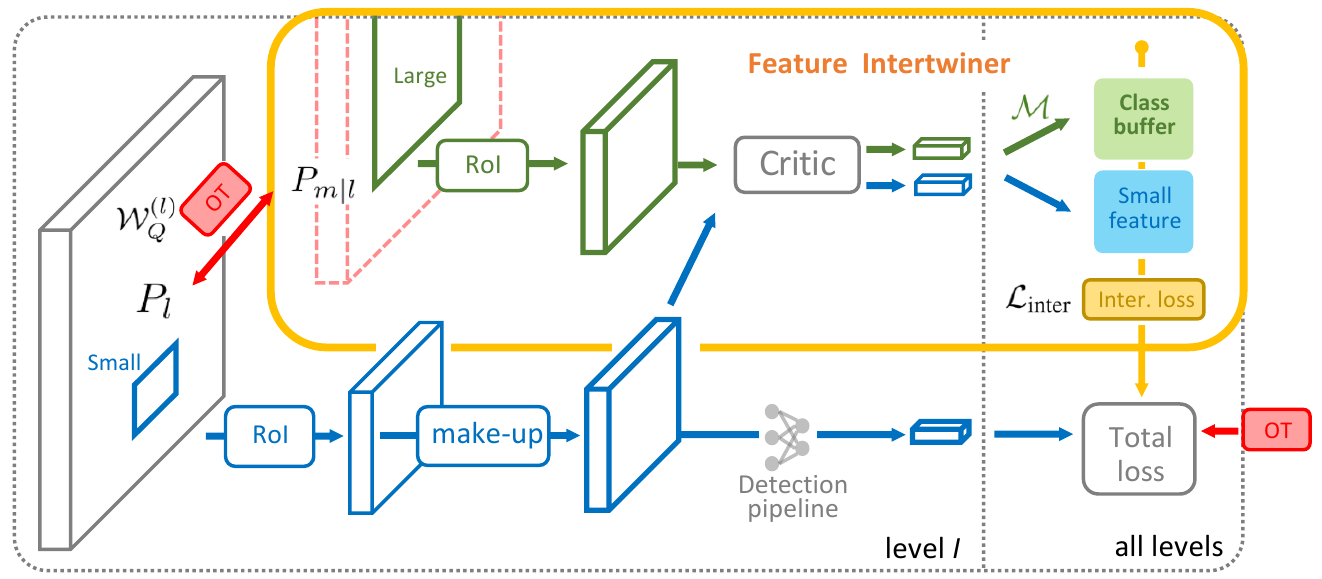}
		\caption{
			Feature intertwinter overview. 
			Blue blobs stands for the less reliable set (small objects) and green for the reliable set (large ones).
			For current level $l$, 
			feature map $P_l$ of the small set is first passed into a RoI-pooling layer. Then it is fed into a make-up layer, which fuels back the information lost during RoI; it is optimized via the intertwiner module  (yellow rectangle), with aid of the reliable set (green). `OT' (in red) stands for the optimal transport divergence, which aligns information between levels (for details see Sec. \ref{sec:optimal-transport-divergence-as-information-alignment}).
			$P_{m | l}$ is the input feature map of the reliable set for the RoI layer; $m$ indicates higher level(s) than $l$.
		} \label{fig:overview}
\end{figure}

\subsection{Feature Intertwiner Overview}\label{sec:turbo-boost-module}

We now explicitly depict how the idea of feature intertwiner could be adapted into the object detection framework.
Fig. \ref{fig:overview} describes the overall pipeline of the proposed InterNet.

A network is divided into several levels 
based on the spatial size of feature maps. For each level $l$, 
we split the set of region proposals 
into two categories: one is the large-region set whose size is larger than the output size of RoI-pooling layer and another the small-region set whose size is smaller. These two sets corresponds to the reliable and less reliable sets, respectively. For details on the generation of these two sets in object detection, 
refer to Sec. \ref{sec:proposal-assignment} in the appendix.
%
%
Feature map $P_l$ at level $l$ is fed into the RoI layer and then passed onto a \textit{make-up} layer. This layer is designed to fuel back the lost information during RoI and compensate necessary details for instances of small resolution.
The refined high-level semantics after this layer is robust to factors (such as pose, lighting, appearance, \textit{etc.}) despite sample variations.
It consists of one convolutional layer without altering the spatial size. The make-up unit is learned and optimized via the intertwiner unit, with aid of features from the large object set, which is shown in the upstream (green) of Fig. \ref{fig:overview}.

The feature intertwiner is essentially a data distribution measurement to evaluate divergence between two sets. For the reliable set, the input is directly the outcome of the RoI layer of the large-object feature maps $P_{m|l}$, which correspond to samples of higher level/resolution.
For the less reliable set, the input is the output of the make-up layer.
Both inputs are fed into a \textit{critic} module
to extract further representation of these two sets and provide evidence for intertwiner. 
The critic consists of two convolutions that transfer features to a larger channel size 
and reduce spatial size to one, leaving out of consideration the spatial information.
A simple $l_2$ loss can be used for comparing difference between two sets.
The final loss is a combination of the standard detection losses \citep{ross15_fast_rcnn} and the intertwiner loss across all levels.

The detailed network structure of the make-up and critic module in the feature intertwiner is shown in the appendix (Sec. \ref{sec:network_feat_inter}). There are two problems when applying the aforementioned pipeline into application. The first is that the two sets for the same category often do not occur simultaneously in one mini-batch; the second is how to choose the input source for the reliable set, \textit{i.e.}, feature map $P_{m | l}$ for the large object set. We address these two points in the following sections.

\subsection{Class Buffer}
The goal of the feature intertwiner is to have samples from less reliable set close to the samples \emph{within the same category} from the reliable set.
In one mini-batch, however, it often happens that samples from the less reliable set are existent while samples of the same category from the reliable set are non-existent (or vice versa). This makes it difficult to calculate the intertwiner loss between two sets.
To address this problem, we use a  {buffer} $\mathcal{B}$
to store the \textit{representative} (prototype) for each category. Basically the representative is the mean feature representation  from large instances.

Let the set of features from the large-region object on all levels be $\mathbf{F}_{\text{critic}}^{(\texttt{large})}$;
each sample consisting of the large set $\mathbf{F}$ be $\bm{f}^{(j)}$, where $j$ is the sample index and its feature dimension is $d$.
The buffer could be generated as a mapping from sample features to class representative:
\begin{gather}
\mathcal{B} = [\bm{b}_1, \dots, \bm{b}_i, \dots, \bm{b}_{N_{\text{cls}}}] = \mathcal{M}  \bigg[ \mathbf{F}_{\text{critic}}^{(\texttt{large}, 1)}, \ldots, \mathbf{F}_{\text{critic}}^{(\texttt{large}, l)}, \ldots, \mathbf{F}_{\text{critic}}^{(\texttt{large}, L)} \bigg], \\
%
{
\bm{b}_{i^*} = \mathcal{M}\Big[\mathbf{F}_{\text{critic}}^{(\texttt{large}, l)} \Big] =
\frac{1}{Z} \sum_{l, j} \bm{f}_{\text{critic}}^{(\texttt{large}, l, j)},~~~\text{where}~\mathbf{F}_{\text{critic}}^{(\texttt{large}, l)}=\{\bm{f}_{\text{critic}}^{(\texttt{large}, l, j)}\in \mathbb{R}^{d} \}, \label{representative}
}
\end{gather}
where the total number of classes is denoted as $N_{\text{cls}}$.
Each entry $\bm{b}_i$ in the buffer $\mathcal{B}$ is referred to as the representative of class $i$. Every sample, indexed by $j$ in the large object set, contributes to the class representative $i^*$ if its label belongs to $i^*$.
Here we denote $i^*$ as the label of sample $j$;
%
and $Z$ in Eqn. (\ref{representative}) denotes the total number of instances whose label is $i^*$.
The representative is deemed as a reliable source of feature representation and could be used to guide the learning of the less reliable set.
There are many options to design the mapping $\mathcal{M}$, \textit{e.g.}, the weighted average of 
	all features in the past iterations during training within the class 
as shown in Eqn. (\ref{representative}), feature statistics from only a period of past iterations, \textit{etc}. We empirically 
discuss different options in Table \ref{tab:ablation:buffer design}. 

Equipped with the class buffer, we define the intertwiner loss between two sets as:
\begin{equation}
\mathcal{L}_{\text{inter}} = \sum_{l, j} \mathcal{D} \big(  \bm{f}_{\text{critic}}^{(\texttt{small}, l, j)}, \mathcal{B} \big),
\end{equation}
where $\mathcal{D} $ is a divergence measurement;
$\bm{f}_{\text{critic}}^{(\texttt{small}, l, j)}$ denotes the semantic feature after critic of the $j$-th sample at level $l$ in the less reliable set (small instances).
Note that the feature intertwiner is proposed to
optimize the feature learning of the less reliable set for each level. During inference, the green flow as shown in Fig. \ref{fig:overview} for obtaining the class buffer will be removed.

\textbf{Discussion on the intertwiner.} \textbf{(a)} Through such a mutual learning, features for small-region objects gradually encode the affluent details from large-region counterparts, ensuring that the semantic features within one category should be as much similar as possible despite the visual appearance variation caused by resolution change. 
The resolution imperfection of small instances inherited from the RoI interpolation is compensated by mimicking a more reliable set. 
Such a mechanism could be seen as a teacher-student guidance in the self-supervised domain \citep{chen2017_obj_det_data_dis}.
\textbf{(b)} 
It is observed that if the representative $\bm{b}_i$ is detached in back-propagation process {(\textit{i.e.}, no backward gradient update in buffer)},  performance gets better. 
The buffer is used as the guidance for less reliable samples.
As contents in buffer are changing as training evolves, excluding the buffer from network update would favorably stabilize the model to converge. 
Such a practice shares similar spirit of the replay memory update in deep reinforcement learning.
%
\textbf{(c)} The buffer statistics come from all levels. Note that the concept of ``large'' and ``small'' is a \textit{relative} term: large proposals on current level could be deemed as ``small'' ones on the next level. However, the level-agnostic buffer would always receive semantic features for (strictly) large instances. 
%
%
This is why there are improvements across \textit{all} levels (large or small objects) in the experiments.

\subsection{Choosing Best Feature Map for Large Objects using Optimal Transport }\label{sec:optimal-transport-divergence-as-information-alignment}

How to acquire the input source, 
denoted as $P^{(\texttt{large}, l)}$, \textit{i.e.}, feature maps of large proposals, to be fed into the RoI layer on current level $l$? 
	The feature maps, denoted by $P_l$ or $P_m$, are the output of ResNet at different stages, corresponding to different resolutions. Altogether we use four stages, \textit{i.e.}, $P_2$ to $P_5$; $P_2$ corresponds to feature maps of the highest resolution and $P_5$ has the lowest resolution.
The inputs are crucial since they serve as the guidance targets to be learned by small instances. 
There are several choices, which is depicted in Fig. \ref{fig:OT_design}.

\begin{minipage}{\textwidth}
	\begin{minipage}[b]{0.3\textwidth}
		\centering
		\includegraphics[width=\textwidth]{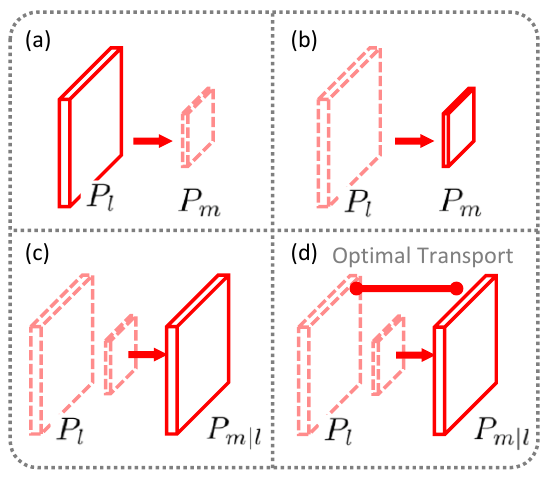}
		\captionof{figure}{Different designs for the input source in the reliable set. Solid shape is the chosen plan in each option. 
		}\label{fig:OT_design}
	\end{minipage}\hfill
	\begin{minipage}[b]{0.68\textwidth}
		\centering
		\tablestyle{2.7pt}{1.01}
		\begin{tabular}{l | l |x{20}x{20}x{20} | x{20}x{20}x{20}}
			\scriptsize option & \scriptsize variant & AP & AP$_{50}$ & AP$_{75}$   & AP$_S$ & AP$_M$ & AP$_L$  \\
			\shline
			\multicolumn{2}{ l |}{ \scriptsize (a) use $P_l$ }   &  \scriptsize 35.1 & \scriptsize 54.9 & \scriptsize 40.7& \scriptsize 20.2 & \scriptsize 38.3 & \scriptsize 48.5\\  \hline
			\multicolumn{2}{ l |}{ \scriptsize (b) use $P_m$ (baseline) } & \cellcolor{Gray} \scriptsize 40.5 & \cellcolor{Gray}\scriptsize 62.8 &\cellcolor{Gray} \scriptsize 47.6 &\cellcolor{Gray} \scriptsize 23.7 & \cellcolor{Gray}\scriptsize 45.2 & \cellcolor{Gray}\scriptsize 53.1\\ \hline
			\scriptsize \multirow{2}{*}{(c)  $P_{m|l}$} &  \scriptsize $\mathcal{F}$ bilinear &  
			\scriptsize 40.6  & \scriptsize  62.9 & \scriptsize 47.6 & \scriptsize 23.9 & \scriptsize 45.4 & \scriptsize 53.1 \\
			& \scriptsize $\mathcal{F}$ neural net*  &  \scriptsize 41.3 & \scriptsize 63.5 & \scriptsize 48.5
			& \scriptsize  24.6& \scriptsize 46.3 & \scriptsize 53.8  \\ \hline
			\multicolumn{2}{r|}{\scriptsize \textit{increase from} (b) \textit{to} (c)*}  & \dt{+0.8} & \dt{+0.7} & \dt{+0.9} & \dt{+0.9} & \dt{+1.1} & \dt{+0.7} \\ \hline
			\scriptsize \multirow{4}{*}{(d) $P_{m|l}$} &  \scriptsize KL, $\mathcal{F}$ neural net 
			&  \scriptsize 41.0  & \scriptsize 63.1& \scriptsize 48.2& \scriptsize 24.5 & \scriptsize 45.7 & \scriptsize 53.4  \\
			&  \scriptsize $l_2$, $\mathcal{F}$ neural net
			&  \scriptsize 41.8  & \scriptsize 64.2& \scriptsize 48.9& \scriptsize 24.7 & \scriptsize 46.0 & \scriptsize 53.8   \\
			&  \scriptsize optimal transport (OT)** &  \scriptsize \cellcolor{Gray}\textbf{42.5 }& \scriptsize \cellcolor{Gray}\textbf{65.1} & \scriptsize\cellcolor{Gray} \textbf{49.4} & \scriptsize\cellcolor{Gray} \textbf{25.4 }& \scriptsize \cellcolor{Gray}\textbf{46.6} & \scriptsize \cellcolor{Gray}\textbf{54.3} \\
			& \scriptsize biased version of OT  &  \scriptsize 42.5 & \scriptsize 65.3 & \scriptsize 48.6& \scriptsize 25.3 & \scriptsize 46.8 & \scriptsize 54.3  \\ \hline
			\multicolumn{2}{r|}{\scriptsize \textit{increase from} (b) \textit{to} (d)**}  & \dt{+2.0} & \dt{+2.3} & \dt{+1.8} & \dt{+1.7} & \dt{+1.4} & \dt{+1.2} \\ 
		\end{tabular}
		\captionof{table}{Numeric results on different input sources for the reliable set (using ResNet-101-FPN model). $\mathcal{F}$ is the up-sampling layer; we use option (d), OT as the final candidate. The biased version of optimal transport is detailed in appendix, Sec. \ref{sec:sinkhorn-divergence}.}\label{tab:OT_design}
	\end{minipage}
\end{minipage}
\medskip

\textbf{Option (a): 
	$P^{(\texttt{large}, l)}=P_l$.}
The most straightforward manner would be using features on current level  as input for large object set. 
This is inappropriate since $P_l$ is trained in RPN specifically for identifying small objects;
adopting it as the source could contain noisy details of small instances.

\textbf{Option (b):{$P^{(\texttt{large}, l)}=P_m$}.} Here $m$ and $l$ denote the index of stage/level in ResNet and  $m>l$. One can utilize the higher level feature map(s), which has the proper resolution for large objects.
{
Compared with $P_l$, $P_m$ have lower resolution and higher semantics. For example, consider the large instances assigned to level $l=2$ (how to assign large and small instances is discussed in the appendix Sec. \ref{sec:proposal-assignment}), $P_m$ indicates three stages $m=3,4,5$.} 
However, among these large instances,
some of them are deemed as {small} objects on higher level $m$ - implying that those feature maps $P_m$ might not carry enough information. They would \textit{also} have to  be up-sampled during the RoI operation for updating the buffer on current level $l$. 
Take Table \ref{tab:anchor_assign} in the appendix for example, among the assigned 98 proposals on level 2, there are 31 (11 on level 3 and 20 on level 4) objects that have insufficient size (smaller than RoI's output). {Hence it  might be inappropriate to directly use the high-level feature map  as well.}
%

\textbf{Option (c): {$P^{(\texttt{large}, l)}=P_{m|l} \triangleq \mathcal{F}(P_m)$}.} $P_m$ is first up-sampled to match the size at $P_l$ and then is RoI-pooled with outcome denoted as $P_{m|l}$. 
The up-sampling operation aims at optimizing a mapping $\mathcal{F}: P_m \mapsto P_{m|l}$ that can recover the information of large objects on a shallow level. $\mathcal{F}$ could be as simple as a bilinear interpolation or  a  neural network.

These three options are empirically reported 
in Table \ref{tab:OT_design}. 
The baseline model in (b) corresponds to the default setting in cases \ref{tab:ablation:buffer design}, \ref{tab:ablation:workflow design} of Table \ref{tab:ablations}, where the feature intertwiner is adopted already.
There is a 0.8\% AP boost from option (b) to (c), suggesting 
that $P_m$ for large objects should be converted back to the feature space of $P_l$. The gain from (a) to (c) is more evident, which verifies that it might not be good to use $P_l$ directly.
 More analysis is provided in the appendix.

%

Option (c) is a better choice for using the reliable feature set of large-region objects. Furthermore, we build on top of this choice and introduce a better alternative to build the connection between $P_l$ and $P_{m|l}$, since the  intertwiner is designed to guide the feature learning of the less reliable set on the current level. 
If some constraint is introduced to keep information better aligned between two sets, the modified input source $P_{m|l}$ for large instance would be more proper for the other set to learn. 

\textbf{Option (d): {$P^{(\texttt{large}, l)}=\texttt{OT}(P_{l}, P_{m|l})$}.} %
The spirit of moving one distribution into another distribution optimally in the most effective manner fits well into the optimal transport (OT) domain \citep{peyre2018_ot_recent_book}. In this work, we incorporate the OT unit between feature map $P_l$ and $P_{m|l}$, which serve as inputs before the RoI-pooling operation.
%
%
%
A discretized version \citep{genevay2017_sinkhorn_loss,cuturi2013_regularized_OT} of the OT divergence is employed as an additional regularization to the loss:
\begin{equation}
\texttt{OT}(P_l, P_{m|l}) \triangleq \mathcal{W}_{Q}(  \mathds{P}_\psi, \mathds{P}_r  )   \xleftarrow[]{\text{discrete}} 
\min_{P \in \mathds{R}_{+}^{ C_2 \times C_1}}
\langle  Q, P\rangle, \label{ot_loss_discrete}
\end{equation}
where the non-positive  $P$ serves as a proxy for the coupling and satisfies $P^\mathsf{T} \mathds{1}_{C_2}  = \mathds{1}_{C_1}, P\mathds{1}_{C_1}  = \mathds{1}_{C_2}$. $ \langle  \cdot, \cdot\rangle$ indicates the Frobenius dot-product for two matrices and $\mathds{1}_m \coloneqq (1/m, \dots, 1/m) \in \mathds{R}_{+}^{m}$.
Now the problem boils down to computing $P$ given some ground cost $Q$.
We adopt the Sinkhorn algorithm \citep{sinkhorn1964_first} in an iterative manner to compute $\mathcal{W}_{Q}$, which is promised to have a differentiable loss function. The OT divergence is hence referred to as Sinkhorn divergence.

Given features maps $P_m$ from higher level, the generator network $\mathcal{F}$ up-samples {them} to match the size of $P_l$ and outputs $P_{m|l}$. The channel dimension of $P_l$ and $P_{m|l}$ is denoted as $C$.
The critic unit $\mathcal{H}$ (not the proposed critic unit in the feature intertwiner) is designed to reduce the spatial dimensionality of input to  a lower dimension $k$ while keeping the channel dimension unchanged. The number of samples in each distribution is $C$. 
The outcome of the critic unit in OT module is denoted as $\bm{p}_l, \bm{p}_{m|l}$, respectively.
We choose cosine distance as the measurement to calculate the distance between 
manifolds. 
The output is known as the ground cost $Q_{x,y}$, where $x,y$ indexes the sample in these two distributions. The complete workflow to compute the Sinkhorn divergence 
is summarized in Alg. \ref{algorithm}. 
%
%
Note that each level owns their own OT module 
$\mathcal{W}_{Q}^{l}(P_l, P_m)=\texttt{OT}(P_{l}, P_{m|l})$. 
The total loss for the detector is summarized as:
\begin{equation}
    \mathcal{L} = \mathcal{L}_{\text{inter}} + \sum_l\mathcal{W}_{Q}^{(l)}(P_l, P_m) +  \mathcal{L}_{\text{standard}},
\end{equation}
where $\mathcal{L}_{\text{standard}}$ is the classification and regression losses defined in most detectors 
\citep{ross15_fast_rcnn}.  

\begin{algorithm}
	\caption{Sinkhorn divergence $\mathcal{W}_{Q} $ 
	adapted for object detection (red rectangle in Fig.\ref{fig:overview})}\label{algorithm}
	\begin{algorithmic}[H]
		\Statex 
		\textbf{Input:}~~~~~~~Feature maps on current and higher levels, $P_l, P_m$ 
		\Statex~~~~~~~~~~~~~~~~~~The generator network $\mathcal{F}$ and the critic unit in OT module $\mathcal{H}$
		\Statex \textbf{Output:}~~~~Sinkhorn loss $\mathcal{W}_{Q}^{l}(P_l, P_m)=\texttt{OT}(P_{l}, P_{m|l})$ 
		\medskip
		\State Upsample via generator $P_{m|l} = \mathcal{F}(P_m)$
		\State Feed both inputs into critic $\bm{p}_l = \mathcal{H}(P_l), \bm{p}_{m|l} = \mathcal{H}(P_{m|l})$ \Comment{\footnotesize $\bm{p}_{(\cdot)}$ size $C\times k$}
		\State $\forall (x,y)$ in $\bm{p}_{l}, \bm{p}_{m|l}$, define the ground cost $Q_{x,y} =\texttt{cosine\_dist}(\bm{p}_{l}, \bm{p}_{m|l})$ \Comment{\footnotesize $Q$ size $C \times C$}
		\medskip
		\Statex Initialize coefficients $b^{(0)} = \mathds{1}_{C}$
		\Statex Compute Gibbs kernel $K_{x,y}=\exp(-Q_{x,y} / \varepsilon)$     \Comment{\footnotesize controlling factor $\varepsilon=0.1$}
		\For  {$l=0$ to $L$}  \Comment{\footnotesize iteration budget $L=10$}
		\State $a^{(l+1)} \coloneqq \frac{ \mathds{1}_{C} }{K^{\mathsf{T}} b^{(l)}}$,  
		$b^{(l+1)} \coloneqq \frac{ \mathds{1}_{C} }{K^{} a^{(l)}}$, \Comment{\footnotesize known as \texttt{Sinkhorn iterate}}
		\EndFor
		\State Compute the proxy matrix $P^{(L)} = \text{diag}(b^{(L)}) \cdot K \cdot \text{diag}(a^{(L)})$
		\State Compute $\mathcal{W}_{Q}$ based on the dot-product in Eqn. (\ref{ot_loss_discrete}): $\langle  Q, P\rangle$.
	\end{algorithmic}
\end{algorithm}

\textbf{Why prefer OT to other alternatives.}    
As proved in \citep{arjovsky2017_wgan}, the OT metric converges while other variants (KL or JS divergence) do not in some scenarios. 
OT provides sensible cost functions when learning distributions supported by low-dim manifolds (in our case, $\bm{p}_l$ and $\bm{p}_{m|l}$) while other alternatives do not.  As verified via experiments in Table \ref{tab:OT_design}, such a property could facilitate the training towards a larger gap between positive and false samples.
In essence, OT metric maps the comparison of two distributions on high-dimensional feature space onto a lower dimension space. 
%
The use of Euclidean distance could improve AP by around 0.5\% (see Table \ref{tab:OT_design}, (d) $l_2$ case), but does not gain as much as OT does. This is probably due to the complexity of feature representations in high-dimension space, especially learned by deep models.


\section{Experimental Results}
We evaluate
InterNet on the 
object detection track of the challenging COCO benchmark \citep{coco}. For training, we follow common practice as in \citep{ren2015_faster_rcnn,he2017_mask_rcnn} and use the \texttt{trainval35k} split (union of 80k images from \texttt{train} and a random 35k subset of images from  40k \texttt{val} split) for training. The lesion and sensitivity studies are reported by
evaluating on the \texttt{minival} split (the remaining 5k images
from \texttt{val}). 
%
For all experiments, we use depth 50 or 101 ResNet \citep{he2016_resnet} with FPN \citep{lin2017_FPN} constructed on top. 
{We base the framework on Mask-RCNN \citep{he2017_mask_rcnn} \textit{without} the segmentation branch. }
All ablative analysis adopt austere settings: training and test image scale only at 512; no multi-scale and data augmentation (except for horizontal flip).
Details on the training and test procedure are provided in the appendix (Sec. \ref{sec:training_test_details}).

\subsection{Ablation study on Intertwiner Module}\label{sec:ablative-analysis}

\textbf{Baseline comparison.} Table \ref{tab:ablation:vs_baseline} lists the comparison of InterNet to baseline, where both methods shares the same setting. 
On average it improves by 2 points in terms of mAP. The gain for small objects is much more evident. Note that our method also enhances the detection of large objects (by 0.8\%), since the last level also participates in the intertwiner update by comparing its similarity feature to the history buffer, which requires features of the same category to be closer to each other. The last level does not contribute to the buffer update though.

{\textbf{Assignment strategy} (analysis based on Sec. \ref{sec:proposal-assignment}).}
Table \ref{tab:ablation:vs_baseline} also investigates the effect of different region proposal allocations. `by RoI size' divides proposals whose area is below the RoI threshold in Table \ref{tab:anchor_assign} as small and above as large; `more on higher' indicates the base value 
in Eqn. (\ref{anchor_ass}) is smaller (=40); the default setting follows \citep{lin2017_FPN} where the base is set to 224. 
Preliminary, we think putting more proposals on higher levels (the first two cases) would balance the workload of the intertwiner; since the default setting leans towards too many proposals on level 2. 
However, there is no gain 
due to the mis-alignment with RPN training. 
The distribution of anchor templates in RPN does not alter accordingly, 
resulting in the inappropriate use of backbone feature maps.

\begin{table*}[t]\vspace{-3mm}
	\subfloat[\textbf{Baseline and proposal assignment strategy}:  intertwiner increases detection of both small and large objects compared to baseline. Putting more proposals on lower level brings more gain.
	\label{tab:ablation:vs_baseline}]
	{\tablestyle{2.5pt}{1.}
		\begin{tabular}{ c | r |x{20}x{20}x{20} | x{20}x{20}x{20}}
			& \scriptsize proposal split & AP & AP$_{50}$ & AP$_{75}$   & AP$_S$ & AP$_M$ & AP$_L$  \\
			\shline
			\scriptsize \multirow{3}{*}{baseline}& \scriptsize by RoI size  &  \scriptsize 30.9 & \scriptsize 53.7 & \scriptsize 35.1& \scriptsize 10.8 & \scriptsize 34.7 & \scriptsize 46.6\\
			& \scriptsize more on higher &  \scriptsize 31.3 & \scriptsize 54.0 & \scriptsize 35.8& \scriptsize 11.4 & \scriptsize 35.1 & \scriptsize 47.5 \\ 
			& \cellcolor{Gray}  \scriptsize default $^{*}$	&  \cellcolor{Gray}\scriptsize 32.8 & \cellcolor{Gray}\scriptsize 55.3 & \cellcolor{Gray}\scriptsize 37.2& \cellcolor{Gray}\scriptsize 12.7 & \cellcolor{Gray}\scriptsize 36.8 & \cellcolor{Gray}\scriptsize 49.3 \\ \hline
			\scriptsize \multirow{3}{*}{intertwiner} &  \scriptsize by RoI size &  \scriptsize 33.7 & \scriptsize 56.1 & \scriptsize 37.6& \scriptsize 13.5 & \scriptsize 37.4 & \scriptsize 50.8 \\
			& \scriptsize more on higher &  \scriptsize 32.3 & \scriptsize 55.7 & \scriptsize 37.1& \scriptsize 12.9 & \scriptsize 36.2 & \scriptsize 49.5  \\ 
			& \cellcolor{Gray}\scriptsize default $^{**}$ & \cellcolor{Gray}\scriptsize \textbf{35.2} & \cellcolor{Gray}\scriptsize \textbf{57.6} &\cellcolor{Gray} \scriptsize \textbf{38.0}& \cellcolor{Gray}\scriptsize \textbf{15.3} &\cellcolor{Gray} \scriptsize \textbf{38.7} & \cellcolor{Gray}\scriptsize \textbf{51.1} \\ \hline
			\multicolumn{2}{r|}{\scriptsize \textit{increase from} $^{*}$ \textit{to}$^{**}$}  & \dt{+2.4} & \dt{+2.1} & \dt{+0.8} & \dt{+2.6} & \dt{+1.9} & \dt{+0.8}
		\end{tabular}
	}
	\hspace{2.5mm}
	\subfloat[\textbf{Feature intertwiner loss}: upper block uses a factor of 1.0.
	 $l_2$ performs slightly better than KL divergence.
	\label{tab:ablation:loss_choice}]{
		\tablestyle{2pt}{1}
		\begin{tabular}{r |x{22}x{22}x{22}}
			& AP & AP$_{50}$ & AP$_{75}$\\
			\shline
			\scriptsize $l_1$ & \scriptsize 34.2 & \scriptsize 57.1 & \scriptsize 37.2\\
			\scriptsize $l_2$ (default) & \cellcolor{Gray} \scriptsize 35.2 & \cellcolor{Gray} \scriptsize\textbf{ 57.6}  &\cellcolor{Gray} \scriptsize \textbf{38.0} \\
			\scriptsize KL \texttt{div} & \scriptsize 34.6 & \scriptsize 57.8 & \scriptsize 37.4 \\ \hline
			\scriptsize $l_1$ (fac=10) & \scriptsize 34.4 & \scriptsize 57.6 & \scriptsize 37.8\\
			\scriptsize $l_2$ (fac=0.1)  & \scriptsize 34.8 &  \scriptsize 58.0 & \scriptsize 37.5 \\
			\scriptsize KL \texttt{div} (fac=10) & \scriptsize \textbf{35.6} & \scriptsize {58.2} & \scriptsize {38.01} \\
			\multicolumn{4}{c}{~}\\
	\end{tabular}}
	\vspace{-1mm}
	\\
	\subfloat[\textbf{Boosted detection feature source}: merging $\bm{f}_{\text{critic}}$ into the detection folllow-up pipeline increases result.
	\label{tab:ablation:turbo_feat_choice}]{
		\tablestyle{1.2pt}{1.}\begin{tabular}{r |x{20}x{20}x{20}}
			& AP & AP$_{50}$ & AP$_{75}$ \\
			\shline
			\scriptsize{separate} & \scriptsize  34.0  & \scriptsize 57.1 & \scriptsize 37.3  \\ \hline
			\scriptsize{naive add} &  \multicolumn{3}{c}{\scriptsize --- fail ---} \\
			\scriptsize{linear} & \cellcolor{Gray}\scriptsize \textbf{35.2 }&  \cellcolor{Gray}\scriptsize\textbf{ 57.6}  & \cellcolor{Gray}\scriptsize \textbf{38.0}    \\
			\multicolumn{4}{c}{~}\\
	\end{tabular}}\hspace{2.2mm}
	%
	%
	\subfloat[\textbf{Buffer choice design} (101-layer): 
	buffer taking in all history  with equal weight ensures best accuracy. Longer size in `partial' block enhances result  and yet possesses more parameters.
	\label{tab:ablation:buffer design}]{
		\tablestyle{1.2pt}{1.}
		\begin{tabular}{c| r | x{22}x{23}x{22}}
			& \scriptsize size/weight	& AP & AP$_{50}$ & AP$_{75}$ \\
			\shline
			\multirow{2}{*}{\scriptsize partial}    	&	\scriptsize 2000	        & \scriptsize 37.3 & \scriptsize 58.5 & \scriptsize 44.7   \\ 
			&	\scriptsize 15k (epoch) & \scriptsize 38.8 & \scriptsize 59.9& \scriptsize  46.1    \\ \hline
			\multirow{2}{*}{\scriptsize all history} 	& \scriptsize	decay weight & \scriptsize 39.2& \scriptsize 60.6& \scriptsize    45.4  \\
			& \scriptsize equal weight   & \scriptsize \cellcolor{Gray}\textbf{40.5} & \cellcolor{Gray}\scriptsize \textbf{62.8} & \cellcolor{Gray}\scriptsize  \textbf{47.6} \\
	\end{tabular}}\hspace{2.2mm}
	%
	%
	\subfloat[\textbf{Workflow design} (101-layer): applying different buffers on each level barely matters; detaching $b_i$ 
	during back-propagation is better.
	\label{tab:ablation:workflow design}]{
		\tablestyle{1.2pt}{1.}
		\begin{tabular}{c| c | x{22}x{22}x{22}}
			& \scriptsize yes?	& AP & AP$_{50}$ & AP$_{75}$ \\
			\shline
			\multirow{2}{*}{\scriptsize multiple $\mathcal{B}$} 	& \scriptsize \cmark	 & \scriptsize 40.58 & \scriptsize 62.83 & \scriptsize 47.62   \\ 
			& \scriptsize \xmark   					& \scriptsize 40.54 & \scriptsize 62.81 & \scriptsize  47.61      \\ \hline
			\multirow{2}{*}{\scriptsize detach $b_i$}    &	\scriptsize 	 \cmark       & \cellcolor{Gray}\scriptsize 40.5 & \cellcolor{Gray}\scriptsize 62.8 & \cellcolor{Gray}\scriptsize  47.6  \\ 
			&	\scriptsize  \xmark & \scriptsize 40.1 & \scriptsize 62.4& \scriptsize 47.3  \\ 
	\end{tabular}}
						\vspace{-.2cm}
	\caption{Ablation study on the component design of feature intertwiner. Gray background is the final default setting adopted in each case. 
		Network is either ResNet-50-FPN or 101.
	}
	\label{tab:ablations}
	\vspace{-.2cm}
\end{table*}

\textbf{Intertwinter loss.} Upper block in Table \ref{tab:ablation:loss_choice} shows a factor of 1.0 to be merged on the total loss whereas lower block depicts a specific factor that achieves better AP than others. The simple  $l_2$ loss achieves slightly better than the KL divergence, where the latter is computed as $L_{\text{inter}} = b \cdot \log(b / \bm{f})$. The $l_1$ option is by around 1 point inferior than these two and yet still verifies the effectiveness of the intertwiner module compared with baseline (34.2 \textit{vs} 32.8) - implying the generalization ability of our method in different loss options.

\textbf{How does the intertwiner module affect learning?} By measuring the  divergence between two sets (\textit{i.e.}, small proposals in the batch and large references in the buffer), we have gradients, as the influence, back-propagated from the critic to make-up layer. In the end, the make-up layer is optimized to enforce raw RoI outputs recovering details even after the loss from reduced resolution. 
The naive design denoted by `separate' achieves 34.0\% AP as shown in Table \ref{tab:ablation:turbo_feat_choice}. 
To further make the influence of the intertwiner stronger, we linearly combine the features after critic with the original detection feature (with equal weights, \textit{aka} 0.5; \textit{not} shown in Fig. \ref{fig:overview}) and feed this new combination into the final detection heads. This improves AP by 1 point (denoted as `linear' in Table \ref{tab:ablation:turbo_feat_choice}). The `naive add' case with equal weights 1 does not work (loss suddenly explodes during training), since the amplitude of features among these two sources vary differently if we simply add them.

\textbf{Does buffer size matter?} Table \ref{tab:ablation:buffer design} shows that it does not. A natural thought could be having a window size of $K$ and sliding the window to keep the most recent features recorded. In general, larger size improves performance (see case `2000' \textit{vs} the size of `one epoch' where batch size is 8, 37.3\% $\rightarrow$ 38.8\%). In these cases,  statistics of large object features for one category cannot reflect the whole training set and it keeps alternating as network is updated. 
Using `all history' data by running averaging not only saves memory  but also has the whole picture of the data. Preliminary, we choose a decayed scheme that weighs more to recent features than ones in the long run, hoping that the model would be optimized better as training evolves. However, experiments does not accord with such an assumption: AP is better where features are equally averaged (\textit{c.f.}, 40.5\% and 39.2\%) in terms of network evolution.

\textbf{Unified or level-based buffer?} Unified. Table \ref{tab:ablation:workflow design} upper block reports such a perspective. In early experiments, we only have one unified buffer in order to let objects on the last level also involved in the intertwiner. Besides, the visual features of large objects should be irrelevant of scale variation. This achieves a satisfying AP already. We also try applying different buffers on each level\footnote{In such case, the last level adopts the buffer on level 2 since it contains the most number of large objects.}. The performance improvement is slight, although the additional memory cost is minor.

\textbf{Other investigations.} As discussed at the end of Sec. \ref{sec:turbo-boost-module}, detaching buffer transaction from gradient update attracts improvement (40.5\% \textit{vs} 40.1\% in Table \ref{tab:ablation:workflow design}). Moreover, we  tried imposing stronger supervision on the similarity feature of large proposals by branching out a cross-entropy loss, for purpose of diversifying the critic outputs among different categories. However, it does not work and this additional loss seems to dominate the training process.

\subsection{Comparison to State-of-the-arts}

\begin{figure}[h]
	\centering
		\includegraphics[width=.85\textwidth]{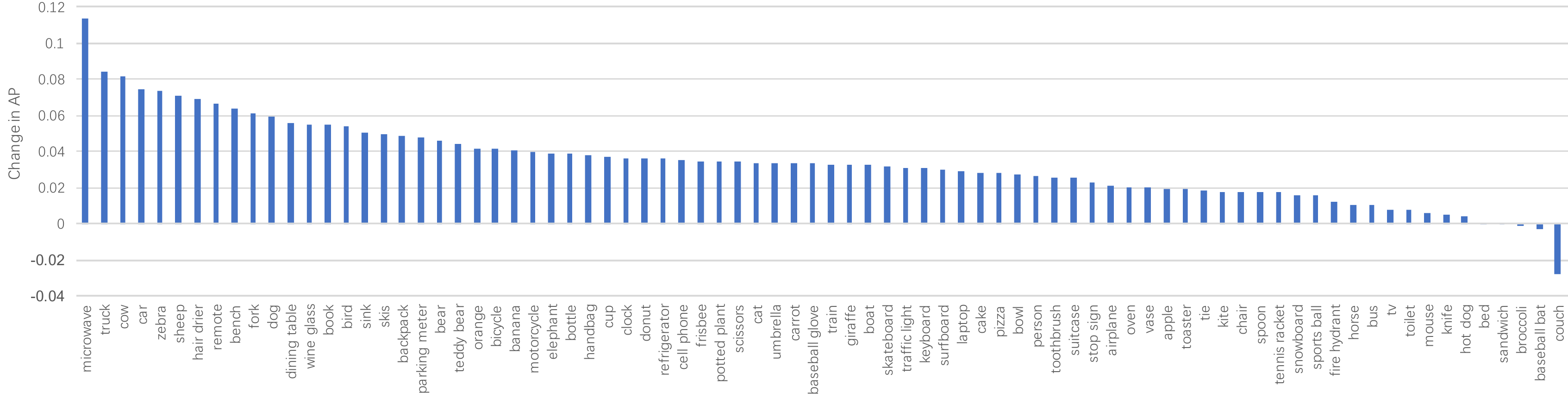}
\vspace{-.2cm}
		\caption{
			Improvement per category after embedding the feature intertwiner on COCO dataset.
		} \label{fig:improve_per_class}
\end{figure}

\textbf{Performance.} We list a comparison of our InterNet with previous state-of-the-arts in Table \ref{tab:final_compare_complete} in the appendix. Without multi-scale technique, ours (42.5\%) still favorably outperforms other two-stage detectors (\textit{e.g.}, Mask-RCNN, 39.2\%) as well as one-stage detector (SSD, 31.2\%).
Moreover, we showcase in Fig. \ref{fig:improve_per_class} the per-class improvement between the baseline and the improved model 
after adopting feature intertwiner in Table \ref{tab:ablation:vs_baseline} (two gray rows). The most improved classes are ‘microwave’, ‘truck’ while the results in ‘couch’, ‘bat’ decrease. Most small-size categories get improved.
{
As for the distinct drop for the `couch' class, we find that for a large couch among samples on COCO, usually there sit a bunch of people, stuff, pets, \textit{etc}. And yet the annotations in these cases would cover the whole scenario including these “noises”, making the feature representation of the large couch quite inaccurate. The less accurate features would guide the learning of their small counterparts, resulting in a lower AP for this class.}

\textbf{Model complexity and timing.} The feature intertwiner only increases three light-weight conv. layers at the make-up and critic units. The usage of class buffer could take up a few GPU memory on-the-fly; however, since we adopt an `all-history' strategy, the window size is just 1 instead of a much larger $K$. 
The additional cost to the overall model parameters is also from the OT module for each level; however, we find using just one conv. layer for the critic $\mathcal{H}$ and two conv. layers with small kernels for generator $\mathcal{F}$ is enough to achieve good result. Training on 8 GPUs with batch size of 8 takes around 3.4 days; this is slower than Mask-RCNN reported in \citep{he2017_mask_rcnn}. The memory cost on each card is 9.6 GB, compared with baseline 8.3 GB. 
The inference runs at 325ms per image (input size is 800) on a Titan Pascal X, increasing around 5\% time compared to baseline (308 ms). 
We do {not} intentionally optimize the codebase, however.

\section{Conclusion and Future Work}

In this paper, we propose a feature intertwiner module to 
leverage the features from a more reliable set to help guide the feature learning of another less reliable set. This is a better solution for generating a more compact centroid representation in the high-dimensional space. It is assumed that the high-level semantic features within the same category should resemble as much as possible among samples with different visual variations. The mutual learning process helps two sets to have closer distance within the cluster in each class. The intertwiner is applied on the object detection task, where a historical buffer is proposed to address the sample missing problem during one mini-batch and the optimal transport (OT) theory is introduced to enforce the similarity among the two sets. Since the features in the reliable set serve as teacher in the feature learning, careful preparation of such features is required so that they would match the information in the small-object set. This is why we design different options for the large set and finally choose OT as a solution.
With aid of the feature intertwiner, we improve the detection performance by a large margin compared to previous state-of-the-arts, especially for small instances. 

{
Feature intertwiner is positioned as a general alternative to feature learning.
As long as there exists proper division of one reliable set and the other less reliable set, one can apply the idea of utilizing the reliable set guide the feature learning of another, based on the hypothesis that these two sets share similar distribution in some feature space. One direction in the future work would be applying feature intertwiner into other domains, \textit{e.g.}, data classification, if proper set division are available.
}

\subsubsection*{Acknowledgments}
We thank Buyu Li for helpful comments in a preliminary version of this work. H. Li and S. Shi are supported by the Hong Kong PhD Fellowship Scheme. This project is also supported by 
the
Research Grants Council of Hong Kong under grant 
CUHK14208417, CUHK14202217, and the Hong Kong Innovation and Technology Support Programme Grant ITS/121/15FX.

\bibliography{bib_library/deep_learning,bib_library/my_pub,bib_library/misc,bib_library/obj_det}
\bibliographystyle{iclr2019_conference}

\section{Appendix}

\subsection{More related work}
\textbf{Self-supervised learning.} The buffer in the feature intertwiner can be seen as utilizing non-visual domain knowledge on a set of data to help 
supervise the feature learning for another set 
in high-dimensional space. Such a spirit falls into the self-supervised learning domain.
In \citep{chen2017_obj_det_data_dis}, Chen \textit{et al.} proposed a knowledge distillation framework to learn compact and accurate object
detectors. A teacher model with more capacity is designed to provide strong information and guide the learning of a lite-weight student model. 
%
The center loss \citep{wen2016_center_loss} is formulated to learn a class center and penalize 
samples that have a larger distance with the centroid. 
It aims at enlarging inter-class resemblance with cross-entropy (CE) loss as well as narrowing down  inner-class divergence for face recognition. 
%
In our work,
the feature intertwiner gradually aggregates statistics of a meta-subset and utilizes them as targets
during the feature learning of a less accurate (yet holding a majority) subset. 
We are inspired by the proposal-split mechanism in object detection domain to learn recognition at separate scales in the network.
{
The \textbf{self-paced learning} framework \citep{kumar2010_self_paced} deals with two sets as well, where the easy examples are first introduced to optimize the hidden variable and later on during training, the hard examples are involved. There is no interaction between the two sets. The division is based on splitting different samples. In our framework, the two sets mutually help and interact with each other. The goal is towards optimizing a more compact class centroid in the feature space. 
These are two different branches of work.
}

\textbf{Optimal transport (OT)} has been applied in two important tasks. One is for transfer learning in the domain adaption problem. Lu \textit{et al.} \citep{lu2017ot_transfer_learn} explored prior knowledge in the cost matrix and applied OT loss as a soft penalty for bridging the gap between target and source predictions. Another is for estimating generative models. In \citep{salimans2018improving}, a metric combined with OT in primal form with an energy distance results in a highly discriminative feature representation with unbiased gradients. Genevay \textit{et al.} \citep{genevay2017_sinkhorn_loss} presents the first tractable method to train large-scale generative models using an OT-based loss. We are inspired by these works in sense that OT metric is favorably competitive 
to measure the divergence between two distributions supported on low-dimensional manifolds.

\subsection{Assignment of large and small sets in object detection}\label{sec:proposal-assignment}
In this paper we adopt the ResNet model \citep{he2016_resnet} with feature pyramid dressings \citep{lin2017_FPN} 
constructed on top. 
It generates five levels
of feature maps to serve as inputs for the subsequent RPN and detection branches. 
Denote the level index as $l=\{1, \dots, 5\}$ and the corresponding feature maps as $P_l$. Level $l=1$ is the most shallow stage with more local details for detecting tiny objects and level $l=5$ is the deepest stage with high-level semantics. 


Let $\mathcal{A}=\{a_j\}$ denote the whole set of proposals generated by RPN from $l_2$ to $l_6$ (level six is generated from $l_5$, for details refer to \citep{lin2017_FPN}).
%
The region proposals are divided into different levels from $l_2$ to $l_5$:
\begin{equation}
a_j^{(l)} \rightarrow l = a_0+\log ( \sqrt{\text{Area}(a_j)} / \texttt{base} ), \label{anchor_ass}
\end{equation}
where 
$a_0$=4 as in \citep{lin2017_FPN}; $\texttt{base}$=224 is the canonical ImageNet pre-training setting.

Table \ref{tab:anchor_assign} shows a detailed breakdown\footnote{Each sample has 200 proposals with input size being 512. Batch size is 2, resulting in 400 proposals in total. Statistics are \textit{averaged per iteration}, based on the output of RPN network during training.} of the proposal allocation based on Eqn. (\ref{anchor_ass}). We can see most proposals from RPN focus on identifying small objects and hence are allocated at shallow level $l=2$. The threshold is set to be the ratio of RoI output's area over the area of feature map.
For example, threshold on $l=3$ is obtained by $(14/64)^2$, where 14 is the RoI output size as default setting.
Proposals whose area is \textit{below} the threshold suffer from the inherent design during RoI operation - these feature outputs are up-sampled by a simple interpolation.
The information of small regions is already lost and RoI layer does not help much to recover them back. 
As is shown on the fourth row (``below \# / above \#''), such a case holds the majority. This observation brings in the necessity of designing a meta-learner to provide guidance on feature learning of small objects due to the loophole during the RoI layer.

\begin{table*}[h]
	\tablestyle{8pt}{1.01}
	\begin{tabular}{c  | x{40} x{30} x{30} x{30} }
		level	$l$ & 2   & 3 & 4 & 5 \\
		\shline 
		proposal \# \scriptsize (perc.)    & 302 \scriptsize (75\%)     & 36 \scriptsize (9\%)& 54 \scriptsize (14\%)& 8 \scriptsize (2\%) \\ \hline
		threshold  & \scriptsize 0.012 & \scriptsize 0.0479 & \scriptsize 0.1914  & \scriptsize 0.7657 \\ 
		below \# / above \#  & 263 / 39 & 25 / 11 & 34 / 20  & 8 / 0\\ \hline
		intertwiner small \# & 302 & 36 & 54 & 8  \\
		intertwiner large \# & 98 
		& 62 
		& 8 
		& -
	\end{tabular}
	\caption{Proposal assignment on each level before RoI operation. `below \#' indicates how many proposals are there whose size is below the size of RoI output.
		`intertwiner large \#' stands for how many proposals are used for supervising the learning of small objects.}
	\label{tab:anchor_assign}
\end{table*}

For  level $l$ in the network, we define \texttt{small} proposals (or RoIs) to be those already assigned by (\ref{anchor_ass}) and \texttt{large} to be those above $l$:
\vspace{-.1cm}
\begin{equation}
a^{(l, \texttt{s})} \leftarrow a_j^{(l)},~~~~a^{(l, \texttt{b})} = \bigcup_{m>l} a_j^{(m)}, \label{assign}
\end{equation}
where the superscript $\texttt{s,b}$ denotes the set of small and large proposals, respectively.
The last two rows in Table \ref{tab:anchor_assign} show an example of the assignment.
These RoIs are then fed into the RoI-pooling layer\footnote{In this paper, we opt for the RoIAlign \citep{he2017_mask_rcnn} option in the RoI layer; one can resort to other options nonetheless. We use term RoI layer, RoI-pooling layer, RoI operation, to refer to the same process.} to generate output features maps for the subsequent detection pipeline to process. 

One may wonder the last level do not have large objects for reference based on Eqn. (\ref{assign}). 
%
In preliminary experiments, leaving proposals on the last level out of the intertwiner could already improve the overall performance; however, if the last level is also involved (since the buffer is shared across all levels), AP for large objects also improves. 
See the experiments in Sec. \ref{sec:ablative-analysis} for detailed analysis. 

\subsection{Sinkhorn divergence}\label{sec:sinkhorn-divergence}

Let $u', u$ indicate the individual sample after degenerating high-dimensional features $P_{m|l}, P_l$ from two spaces into low manifolds.
$u', u$ are vectors of dimension $k$. The number of samples in these two distributions is denoted by $C_1$ and $C_2$, respectively.
The OT metric between two joint probability distributions supported
on two  spaces $(\mathcal{U}, \mathcal{U})$ is defined as the solution of the linear program \citep{cuturi2013_regularized_OT}. Denote the data and reference distribution as
$\mathds{P}_\psi, \mathds{P}_r \in \text{Prob}(\mathcal{U})$\footnote{$\text{Prob}(\mathcal{U})$ is the set of probability distributions over a metric space $\mathcal{U}$.}, respectively, we have the continuous form of OT divergence: 
\begin{equation}
\mathcal{W}_{Q}(  \mathds{P}_\psi, \mathds{P}_r  )= 
\inf_{\gamma \in \Gamma(\mathds{P}_\psi, \mathds{P}_r   )}
\mathds{E} \bigg[ \int_{    
	\mathcal{U} \times \mathcal{U}
} Q(u', u) d \gamma(u', u)  \bigg], \label{ot_loss} \\
\end{equation}
where $\gamma$ is a coupling; $\Gamma$ is the set of couplings that consists of joint distributions.

Intuitively, $\gamma(u', u)$ implies how much ``mass'' must be transported from $u'$ to $u$ in order to transform the distribution $\mathds{P}_\psi$ into  $\mathds{P}_r$; $Q$ is the ``ground cost" to move a unit mass. 
Eqn. (\ref{ot_loss}) above becomes the \textit{p}-Wasserstein distance (or loss, divergence) between probability measures when $\mathcal{U}$ is equipped with a distance $\mathcal{D}_{\mathcal{U}}$ and $Q=\mathcal{D}_{\mathcal{U}}(u', u)^p$, for some exponent $p$.

The biased version of Sinkhorn divergence used in Table \ref{tab:OT_design} is defined by:
\begin{equation*}
2\mathcal{W}_{Q}(  \mathds{P}_\psi, \mathds{P}_r  ) - \mathcal{W}_{Q}(  \mathds{P}_r, \mathds{P}_r  ) - \mathcal{W}_{Q}(  \mathds{P}_\psi, \mathds{P}_\psi  ).  
\nonumber
\end{equation*}

\textbf{More analysis on Table \ref{tab:OT_design}}.
%
All these options have been discussed explicitly at the beginning of Sec. \ref{sec:optimal-transport-divergence-as-information-alignment}.
Option (a) is inferior due to the inappropriateness of feature maps; (b) serves as the baseline and used as the default setting in Table \ref{tab:ablations}.
Options in (c) verifies that up-sampling feature maps from higher-level onto current level is preferable; $\mathcal{F}$ being a neural net ensures better improvement.
Options in (d) illustrates the case where a supervision signal is imposed onto pair $(P_l, P_{m|l})$ to make better alignment between them.
We can observe that OT outperforms other variants in this setup. Moreover, we tried a biased version \citep{genevay2017_sinkhorn_loss} of the Sinkhorn divergence.
However, it does not bring in much gain compared to the previous setup. Besides, it could burden system efficiency during training (although it is minor considering the total time per iteration). Such a phenomenon could result from an improper update of critic and generator inside the OT module, since the gradient flow would be iterated twice more for the last two terms above.


\textbf{Extending OT divergence to image classification.} We also testify OT divergence on CIFAR-10 \citep{cifar} where feature maps between stages 
are aligned via OT. 
Test error decreases by around 1.3\%. This suggests the potential application of OT 
in various vision tasks.
Different from OT in generative models, we deem the channel dimension as different samples to compare, instead of batch-wise manner as in \citep{salimans2018improving}; and treat the optimization of $\mathcal{F}$ and $\mathcal{H}$ in a unified $\min$ problem, as opposed to the adversarial training \citep{genevay2017_sinkhorn_loss}.

\subsection{Comparison to state-of-the-arts on COCO and PASCAL VOC}

Table \ref{tab:final_compare_complete} reports the performance of our model compared with other state-of-the-arts on COCO dataset. We can observe that it outperforms all previous one-stage or two-stage detectors by a large margin. 
The multi-scale technique bundled with data augmentation increases detection accuracy 
in a more evident manner, which is commonly adopted in most detectors. 
{
The updated result in Mask-RCNN is reported as well. It increases the original performance from 38.2\% to 43.5\%
by switching the backbone structure to ResNetX, an updated baseline model, ImageNet-5k pre-training and train-time augmentation. It is better than ours (42.5\% without multi-scale version). This is probably mainly due to the change of network structure. Our multi-scale version (44.2\%) is better than the updated Mask-RCNN result, however.}

\begin{table*}[h]
	\tablestyle{0.5pt}{1.05}
	\centering
	\begin{tabular}{   r  |c|x{22}x{22}x{22}|x{22}x{22}x{22}}
		& backbone
		& AP & AP$_{50}$ & AP$_{75}$
		& AP$_S$ & AP$_M$ &  AP$_L$\\ [.1em]
		\shline
		\emph{One-stage detector} & & & & & & & \\
		~YOLOv2 \citep{redmon2016_yolo_v2} & DarkNet-19 & 21.6 & 44.0 & 19.2 & 5.0 & 22.4 & 35.5 \\
		~SSD513 \citep{liu2015_ssd} & ResNet-101-SSD & 31.2 & 50.4 & 33.3 & 10.2 & 34.5 & 49.8 \\
		~DSSD513 \citep{fu2017_dssd} & ResNet-101-DSSD & 33.2 & 53.3 & 35.2 & 13.0 & 35.4 & 51.1 \\
		\hline
		\emph{Two-stage detector} & & & & & & & \\
		~F-R-CNN+++ \citep{he2016_resnet} & ResNet-101-C4 & 34.9 & 55.7 & 37.4 & 15.6 & 38.7 & 50.9\\
		~F-R-CNN w FPN \citep{lin2017_FPN} & ResNet-101-FPN
		& 36.2 & 59.1 & 39.0 & 18.2 & 39.0 & 48.2\\
		~F-R-CNN by G-RMI \citep{huang2017_speed_accuracy} & Incept.-ResNet-v2 
		& 34.7 & 55.5 & 36.7 & 13.5 & 38.1 & 52.0\\
		~F-R-CNN w TDM \citep{shrivastava2016_top_down_modulation} & Incept.-ResNet-v2-TDM
		& 36.8 & 57.7 & 39.2 & 16.2 & 39.8 & {52.1}\\
		~R-FCN \citep{dai2016_rfcn} & ResNet-101 & 29.9 & 51.9  & - & 10.8 & 32.8 & 45.0 \\
		Mask RCNN \citep{he2017_mask_rcnn} & ResNet-101-FPN & 38.2 & 60.3 & 41.7 & 20.1 & 41.1 & 50.2 \\
		~{RetinaNet} \citep{lin2017_focal_loss} & ResNet-101-FPN & 39.1 & 59.1 & 42.3 & 21.8 & 42.7 & 50.2 \\
	Mask RCNN, updated in \citep{he2017_mask_rcnn}  & ResNetX-101-FPN 
	& 43.5 &  65.9 &  47.2 & - & - & - \\
		\hline 
		~\textbf{InterNet} (ours) & ResNet-101-FPN & 42.5 & 65.1&  49.4& 25.4 & 46.6 & 54.3 \\
		~\textbf{InterNet
		} (ours) \scriptsize multi-scale & ResNet-101-FPN &  \textbf{44.2}& \textbf{67.5}&  \textbf{51.1}&  \textbf{27.2}&  \textbf{50.3}& \textbf{57.7} \\
	\end{tabular}
	\caption{
		Object detection \textit{single-model} performance (bounding box AP) on the COCO \texttt{test-dev}. 
		We show two versions of InterNet that incorporates both the feature intertwiner module and OT agreement. The latter is achieved with data augmentation, 1.5$\times$ longer training time and multi-scale training.  `F-R-CNN' stands for Faster R-CNN.
		Our InterNet is also a two-stage detector.
	}\label{tab:final_compare_complete}
\end{table*}

\begin{table*}[h]
	\centering
	\begin{tabular}{   r |  l l l}
		Model & Structure & Training data & mAP \\ \toprule
		Fast R-CNN \citep{ross15_fast_rcnn} & VGG-16 & 07 & 66.9 \\
		Faster R-CNN \citep{he2016_resnet} & VGG-16 & 07 & 69.9 \\
		SSD512 \citep{liu2015_ssd} & VGG-16 & 07 & 71.6 \\
		InterNet (ours) & VGG-16 & 07 & 73.1 \\ 
		\midrule
		Faster R-CNN \citep{he2016_resnet} & ResNet-101 & 07+12 & 76.4 \\
		R-FCN \citep{dai2016_rfcn} & ResNet-101 & 07+12 & 80.5 \\
		InterNet (ours) & ResNet-101 & 07+12 & 82.7 \\
	\end{tabular}
	\vspace{-.2cm}
	\caption{
		{Comparison of our model with feature intertwiner to other methods on PASCAL VOC 2007 test set. Here we adopt two backbone options: ResNet-101 and VGG-16 without FPN to fairly compare with others. The number of levels is 4, the same as on COCO benchmark.}
	}\label{tab:pascal_voc}
\end{table*}

{To 
further verify the effectiveness of the feature intertwiner, we further 
conduct experiments on the PASCAL VOC 2007 dataset. The results are shown in Table \ref{tab:pascal_voc}.
Two network structures are adopted. For ResNet-101, the division of the four levels are similar as ResNet-101-FPN on COCO; for VGG-16, we take the division similarly as stated in SSD \citep{liu2015_ssd}. Specifically, the output of layer `conv7', `conv8\_2', `conv9\_2' and `conv10\_2' are used for $P_2$ to $P_5$, respectively.
Our method performs favorably against others in both backbone structures on the PASCAL dataset.}

\subsection{Training and test details}\label{sec:training_test_details}
We adopt the stochastic gradient descent as optimizer. Initial learning rate is 0.01 with momentum 0.9 and weight decay 0.0001. Altogether there are 13 epoches for most models where the learning rate is dropped by 90\% at epoch 6 and 10. We find the warm-up strategy \citep{goyal2017_fast_imagenet_train} barely improves the performance and hence do not adopt it. The gradient clip is introduced to prevent training loss to explode in the first few iterations, with maximum gradient norm to be 5. Batch size is set to 8 and the system is running on 8 GPUs.

{
The object detector is based on Mask-RCNN (or Faster-RCNN).
RoIAlign is adopted for better performance. The model is initialized with the corresponding ResNet model pretrained on ImageNet. The new proposed feature intertwiner module is trained from scratch with standard initialization. The basic backbone structure for extracting features is based on FPN network \citep{lin2017_FPN}, where five ResNet blocks are employed with up-sampling layers. The region proposal network consists of one convolutional layer with one classification and regression layer.
The classifier structure is similar as RPN's - one convolution plus one additional classification/regression head.}

Non-maximum suppression (NMS) is used during RPN generation and detection test phase. Threshold for RPN is set to 0.7 while the value is 0.3 during test.
We do not adopt a dense allocation of anchor templates as in some literature \citep{liu2015_ssd,redmon2017_yolo_v1}; each pixel on a level only has the number of anchors the same as the number of aspect ratios (set to 0.5, 1 and 2). Each level $l$ among the five stages owns a unique anchor size: 32, 64, 128, 256, and 512.  
%
{
\subsection{Network structure in feature intertwiner}\label{sec:network_feat_inter}}

{	
The detailed network architecture on the make-up layer and critic layer are shown below.
}

\begin{table*}[h]
	\centering
	\begin{tabular}{   l |  l}
		Output size & Layers in the make-up module \\ \toprule
		$B \times C_l \times 14 \times 14$ & conv2d($C_l, C_l, k=3, \text{padding}=1$) \\
		$B \times C_l \times 14 \times 14$ & batchnorm2d($C_l$) \\
		$B \times C_l \times 14 \times 14$ & relu($\cdot$) \\
	\end{tabular}
	\vspace{-.2cm}
	\caption{
	{
		Network structure of the make-up unit, which consists of one convolutional
layer without altering the spatial size. Input: RoI output of the small-set feature map $P_l$. We denote the output of the make-up layer as $P_l'$.
		$B$ is the batch size in one mini-batch; $C_l$ is the number of channels after the feature extractor in ResNet blocks for each level. For example, when $l=2$, $C_l=256$, \textit{etc}.}
	}\label{tab:make_up_layer}
\end{table*}

\begin{table*}[h]
	\centering
	\begin{tabular}{   l |  l}
		Output size & Layers in the critic module \\ \toprule
		$B \times 512 \times 7 \times 7$ & conv2d($C_l, 512, k=3, \text{padding}=1, \text{stride}=2$) \\
		$B \times 512 \times 7 \times 7$ & batchnorm2d($512$) \\
		$B \times 512 \times 7 \times 7$ & relu($\cdot$) \\
		$B \times 1024 \times 1 \times 1$ & conv2d($512, 1024, k=7$) \\
		$B \times 1024 \times 1 \times 1$ & batchnorm1d($1024$) \\
		$B \times 1024 \times 1 \times 1$ & relu($\cdot$) \\
		$B \times 1024 \times 1 \times 1$ & sigmoid($\cdot$) \\
	\end{tabular}
	\vspace{-.2cm}
	\caption{
	{
		Network structure of the critic unit.
		%
		Input: for large set, it is the RoI output of the large-set feature map $P_{m|l}$ and for small set, it is the output of the make-up layer $P_l'$.
		$B$ is the batch size in one mini-batch; $C_l$ is the number of channels in ResNet blocks.}
	}\label{tab:critic_layer}
\end{table*}

\end{document}